\documentclass{article}
\usepackage[utf8]{inputenc} % allow utf-8 input
\usepackage[backend=biber, style=authoryear]{biblatex}
\usepackage{xcolor,colortbl}
\usepackage{PRIMEarxiv}
\usepackage{geometry}
\usepackage{wrapfig}
\usepackage{subcaption}

\usepackage[T1]{fontenc}    % use 8-bit T1 fonts
\usepackage{hyperref}       % hyperlinks
\usepackage{url}            % simple URL typesetting
\usepackage{booktabs}       % professional-quality tables
\usepackage{amsfonts}       % blackboard math symbols
\usepackage{nicefrac}       % compact symbols for 1/2, etc.
\usepackage{microtype}      % microtypography
\usepackage{lipsum}
\usepackage{fancyhdr}       % header
\usepackage{graphicx}       % graphics
\graphicspath{{media/}}     % organize your images and other figures under media/ folder
\usepackage{tabularx}
\usepackage{float}
\usepackage{amsmath}
\title{Classification of tennis actions using deep learning
}

\author{ 1. Emil Hovad \\
Department of AI and Data Analytics. \\ Alexandra Instituttet A/S, Rued Langgaards Vej 7, 2300 København S\\
\texttt{emil.hovad@alexandra.dk} \\
        Department of Mathematics and Computer Science. Technical University of Denmark.\\ Richard Petersens Plads. Building 324. 2800 Kgs. Lyngby. \\
	\texttt{emilh@dtu.dk} \\
	\And
    2. Therese Hougaard-Jensen \\
	Department of Mathematics and Computer Science. Technical University of Denmark.\\ Richard Petersens Plads. Building 324. 2800 Kgs. Lyngby. \\
	\AND
	3. Line Katrine Harder Clemmensen \\
	 Department of Mathematics and Computer Science. Technical University of Denmark.\\ Richard Petersens Plads. Building 324. 2800 Kgs. Lyngby. \\
	\texttt{lkhc@dtu.dk} \\
}
\begin{document}
\maketitle

%Emil, hvordan hænger introduktion og diskussion sammen? Der står noget om accuracy på 74% F1 i intro og accuracy på 47% F1 ii disukssion. Derudover står der i iintro accuracy på 93% - er det et andet accuracy mål? Eller andet datasæt, det er ikke tydeligt...
% Dbh. Line

\begin{abstract}
Recent advances of deep learning makes it possible to identify specific events in videos with greater precision. This has great relevance in sports like tennis in order to e.g., automatically collect game statistics, or replay actions of specific interest for game strategy or player improvements. In this paper, we investigate the potential and the challenges of using deep learning to classify tennis actions. Three models of different size, all based on the deep learning architecture SlowFast were trained and evaluated on the academic tennis dataset THETIS. The best models achieve a generalization accuracy of 74 \%, demonstrating a good performance for tennis action classification. We provide an error analysis for the best model and pinpoint directions for improvement of tennis datasets in general. We discuss the limitations of the data set, general limitations of current publicly available tennis data-sets, and future steps needed to make progress.
\end{abstract}

% keywords can be removed

\section{Introduction}
% Old school cv
Traditionally, video classification used hand-engineered features to extract information from videos. Popular approaches included extracting spatio-temporal features using first a feature detector and then a feature descriptor. The feature detector localizes and extracts features either densely (\cite{wang2011action}) or at a set of interest points found using dedicated kernels and filters such as the Harris3D detector (\cite{laptev2005space}),  Cuboid (\cite{dollar2005behavior}), and others (\cite{willems2008efficient}). The extracted features were combined into a video-level feature descriptor using, for example, k-means dictionaries, histogram of oriented gradients (HOG) (\cite{dalal2005histograms}), histogram of optical flow (HOF) (\cite{dalal2006human}), SIFT-3D \cite{scovanner20073}, or HOG3D (\cite{klaser2008spatio}) descriptors. Although hand-engineered features can give a fairly good and for many years competitive performance, they are highly problem dependent. 

%Fukushima1980NeocognitronAS
% cnn and deep learning
Convolutional Neural Networks (CNNs) were introduced by (\cite{Fukushima1980NeocognitronAS}) more than four decades ago. Which imitated the visual cortex of humans and are especially suitable for images because they extract local spatial relations. A decade later, CNNs where used to read handwritten postal letters with the LeNet architecture (\cite{le1989handwritten}). However, CNNs did not get much attention in the following years due to the lack of adequate computing power. In the 2000s and thereafter, CNNs had a comeback due to more powerful GPUs and CPUs and larger amounts of training data (\cite{lecun2010convolutional}), such as the ImageNet dataset containing more than 14 million hand-annotated images (\cite{deng2009imagenet}). The latter played a significant role in the advances in image analysis, by hosting an annual ImageNet Challenge for scientist to participate in. In 2012 (\cite{krizhevsky2012imagenet}) proposed AlexNet, a CNN architecture leveraging the improvements in GPU performance. With an error rate more than 10 \% lower than the runner-up in the ImageNet Challenge, AlexNet was a huge breakthrough for CNNs in image analysis. After the succes of AlexNet, CNNs became a dominant method in many computer vision tasks like image classification and object detection. By using deeper networks and new architectures, performance have continued to improve. Good examples of these improvements are networks such as the systematic approach to the layer composition in VGG (\cite{simonyan2014very}), the inception block with different operations type in the same layer in GoogLeNet (\cite{szegedy2015going}) and the skip connection improving the gradient flow for back-propagation in ResNet (\cite{he2016deep}).

% Sports in general
Deep learning approaches have also recently been adopted to sports classification tasks. In (\cite{sozykin2018multi}) they used a three dimensional deep CNN architecture to classify actions in multi-labeled hockey videos. Comparison of a CNN architecture with different combined CNN and LSTM architectures on broadcast videos from the Cricket World Cup was made in (\cite{izadi2018netclips}). Many data-sets for video classification in sports have been proposed throughout the years such as Kinetics 400 (\cite{kay2017kinetics}), Charades (\cite{sigurdsson2016hollywood}),
Olympic (\cite{niebles2010modeling}) and UCF101 (\cite{soomro2012ucf101}) which contain actions spread across multiple different domains. As an example, Kinetics 400 contains classes like bowling, tossing coin, and eating doughnuts. Therefore, much research has been made on these particular sports data-sets, where the focus is on classifying a particular type of sport and not various actions related to a specific sport. To distinguish these classes both motion, object and scenery information will give many more important cues when determining the right class. Sport type classification is highly equipment and environment dependent, which is why the work on these data-sets only peripherally relate to our task. 

% tennis 
In the paper "TenniSet: A Dataset for Dense Fine-Grained Event Recognition, Localisation and Description" by (\cite{faulkner2017tenniset}), the authors used a collection of professional broadcasting videos of tennis matches to perform different tasks within tennis video analysis. 
This included action recognition, detection, and description of tennis videos. For action recognition, a few different deep learning architectures were compared, including a frame-wise classification and different configurations of a CNN-RNN network. All models were trained using raw RGB data, optical flow, or the two combined in a two-stream approach. The best model was found to be the frame-wise two-stream model, although in general, the differences were small, and the best performances were not much better than the RGB frame-wise model. Using only RGB input, the CNN-RNN gave the best results. The authors also found that optical flow alone often had a positive effect no matter the model, which led them to believe that low-level motion information is important for action recognition tasks. 

In one of the first tennis ball tracking papers, Tracknet was proposed (\cite{huang2019tracknet}) and used to track tennis or badminton balls in labelled broadcasting videos. Here the ball was labelled based on pixel position and visibility as four types, 1) no ball, 2) easy, 3) hard and 4) occluded ball. The Tracknet is a neural net which models the ball as a 1-channel grid output with the same resolution as the input frames. The heat-map is modelled as a 2-D normalized Gaussian distribution. The Tracknet was further developed (\cite{rocha2023analysis}) where besides the ball tracking the court lines and the player locations were also detected.
% the link for download does not work.

% nvidia paper (Learning Physically Simulated Tennis Players from Broadcast Videos)
Realistic simulations of tennis players and ball dynamics are accomplished in (\cite{zhang2023vid2player3d}). They used broadcasting videos in combination with simulation to generate realistic movement of tennis characters in a tennis simulation. Their work is based on hierarchical models, which combines a low-level imitation policy with a high-level motion planning policy to control the physics of the tennis character in a motion embedding learned from highly available broadcast videos. This approach can potentially be used to create synthetic tennis data.

% us 
One of the only academic open source video dataset for tennis shots detection is from the paper "THETIS: Three Dimensional Tennis Shots, A human action dataset", (\cite{gourgari2013thetis}). The dataset consists of different tennis actions, like forehand, backhand, backhand slice, etc. They recorded the same 1980 Videos in RGB and added Depth and a Mask, additionally 1217 videos with Skelet2D and Skelet3D were also recorded. In their study, they achieved an average accuracy of 60 \% on the THETIS Depth and an average accuracy
54 \% THETIS Skelet3D and no results were listed for the "pure video data", namely the THETIS RGB. In a later study named "Deep Learning for Domain-Specific Action Recognition in Tennis" by (\cite{vinyes2017deep}), the authors used the THETIS dataset to investigate the challenges within domain-specific action recognition, the task of recognizing actions within a domain rather than telling different domains apart. Their performance for shot type detection was a F1 of 47\% on the THETIS RGB data. In this paper higher accuracy was achieved when splitting the data into amateur players and professionals.

 We choose to use the SlowFast architecture (\cite{feichtenhofer2019slowfast}) as it achieves very high performance on multiple benchmark datasets such as Kinetics 400 (\cite{kay2017kinetics}, Kinetics 600 (\cite{carreira2018short}), Charades (\cite{sigurdsson2016hollywood}), AVA (\cite{gu2018ava}) and we achieve a generalization accuracy of 74 \% on the THETIS RGB data-set. This model uses a two-stream network where each stream is an inflated 2D CNN architecture operating solely on raw video data. Inspired by the THETIS data, several video cameras and markers were applied for recording tennis shots in (\cite{s23052422}). In this data-set the tennis shots where tracked in 3-D with an Attention Temporal Graph Convolutional Network and the highest accuracy was 93 \% for the whole player’s silhouette together with a tennis racket. 

\section{Data}\label{Chap:data}
THETIS consists of a combination of amateur and professional players filmed performing different tennis shots like flat forehand and slice serve. The dataset is designed with great difficulty in mind to facilitate the development of different motion analysis and classification methods.
They recorded the same 1980 Videos in RGB where they added Depth and a Mask. Additionally, 1217 videos with Skelet2D and Skelet3D were recorded.
We use the THETIS RGB of 1980 videos, because of its design towards a wide variety of tennis actions and as it allows for direct comparison with previous work and puts the performance of the used models into an academic perspective using "standard RGB videos" with open-source deep learning models.

\subsection{Tennis video data} \label{sec:tennis_characteristics}
% Describe the characteristics of tennis data. What are the spatial and temporal characteristics?
Tennis is a racket sport with two players playing against each other, positioned on opposite sides of a net. Each player hits a tennis ball with their racket over the net into the other players court. The player who cannot return the ball in a valid way looses the point. Due to the nature of the tennis sport, tennis videos contain 'dense fine-grained events', meaning that much important information is present in very localized spatial and temporal areas of the video. The spatial information of an action is typically connected to the players' movements and the movement of the ball, which takes up only a small part of the total video image. The temporal information of a tennis action is dense because actions are fast and spanning only a few frames. This implies fast change from one action to another, and fast movements both by players and objects. 

%\subsubsection{Tennis is domain-specific data}

\subsection{The THETIS dataset}
\begin{figure}
\centering
    \begin{subfigure}{0.45\columnwidth}
        \centering
    \includegraphics[width=1\textwidth]{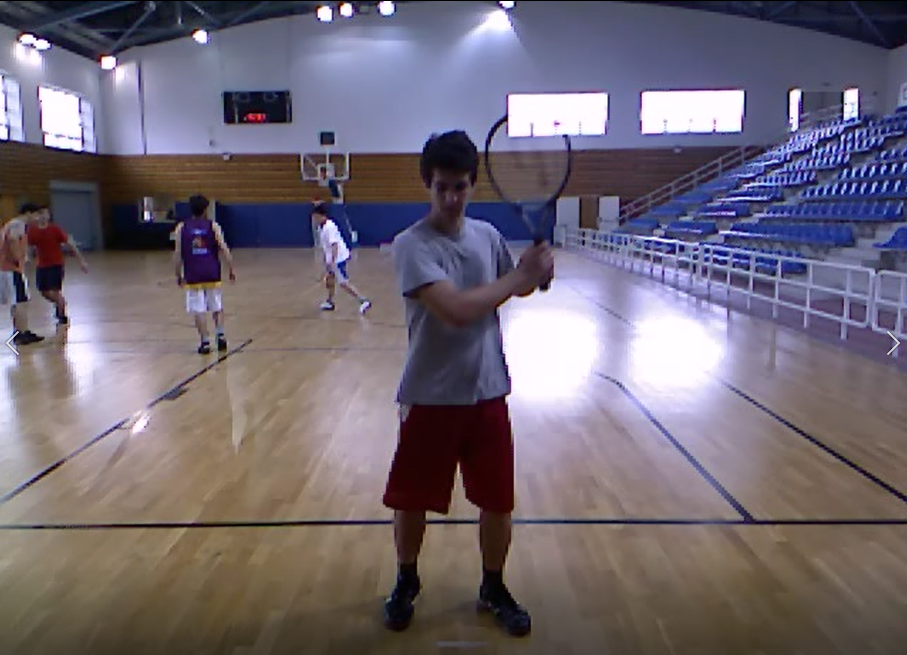} 
        \caption{Example of a basket ball court background \cite{gourgari2013thetis}.}
        \label{fig:background_basket}
    \end{subfigure}
    \begin{subfigure}{0.45\textwidth}
        \centering
        \includegraphics[width=1\textwidth]{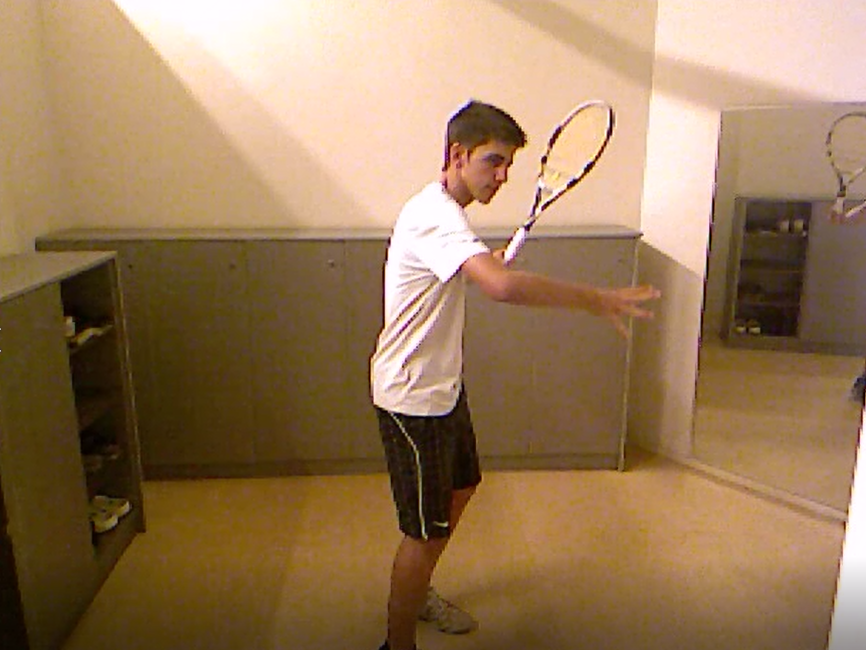}
        \caption{Example of a changing room background \cite{gourgari2013thetis}.}
        \label{fig:background_changing}
    \end{subfigure}
    \caption{The two different backgrounds present in the THETIS dataset}
    \label{fig:thetis_backgrounds}
\end{figure}

The THETIS dataset is a scientific dataset proposed in (\cite{gourgari2013thetis}). It consist of 1980 videos showing 55 different tennis players performing 12 different tennis shots. Each player performs each shot three times resulting in 165 videos in each class. The players' level of tennis experience range from beginner to intermediate. The players are filmed facing the camera, in close proximity to the camera. They perform the tennis shots using a racquet, but no ball. Therefore, the shots are limited to the movements of the player and do not include the resulting movement of the ball, which is often part of the characteristics of a shot as described in Section \ref{sec:tennis_characteristics}. The players stand in front of one of two different backgrounds, see Figure \ref{fig:thetis_backgrounds}. None of the backgrounds are completely static. One of them is in a changing room, where the player is sometimes reflected in the mirror and the other is on a basketball court where a varying number of people are walking around or playing basketball in the background.

\subsection{Classes}
The dataset contains 12 classes with some of the most common types of shots in tennis. The classes are sub-classes of tennis shots, meaning different types of serves, backhands, and forehands are present in the dataset. The classes are shown in Table \ref{tab:thetis_classes}.

\begin{table}
    \centering
    \caption{Classes of the THETIS dataset}
    \begin{tabular}{l} \hline
         \textbf{Class} \\ \hline
         backhand \\
         backhand 2 hands \\
         backhand slice \\
         backhand volley \\
         flat service \\
         forehand flat \\
         forehand open-stands \\
         forehand slice \\
         forehand volley \\
         kick service \\
         slice service \\
         smash \\ \hline
    \end{tabular}
    \label{tab:thetis_classes}
\end{table}

\subsubsection{Statistics}
\begin{figure}
    \centering
    \includegraphics[width=0.9\linewidth]{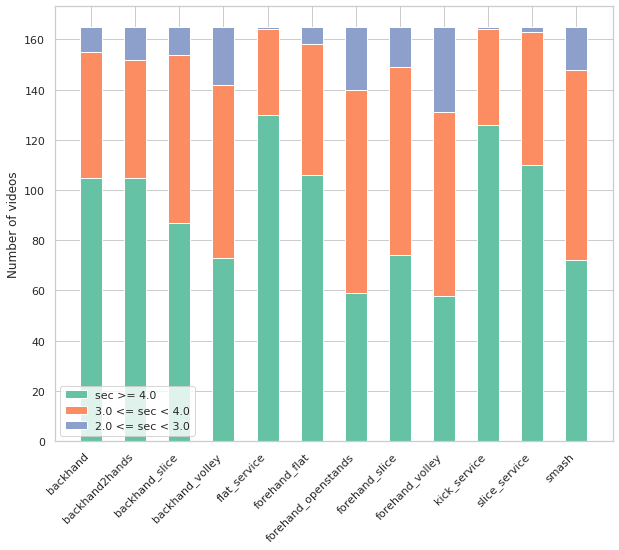}
    \caption{Number of videos per class in the THETIS dataset arranged by length of the videos}
    \label{fig:thetis_class_length}
\end{figure}

The classes in THETIS are evenly distributed with $165$ videos in each class. The length of the videos vary slightly from 2 to 5 seconds as shown in Figure \ref{fig:thetis_class_length}.

\subsubsection{Train, val, test split}
The dataset is split into training, validation, and test set. The dataset is split in 70 \% for training, 20 \% for validation, and 10 \% for testing. Due to the small size of the dataset, only $10 \%$ of the videos were used for the test set such that the model could be trained on more data. 

% Describe the splits of the datasets
%\subsubsection{Train, val, test split}
%The dataset is split into training, validation, and test set in a stratified manner. The dataset is split into 60 \% for the training dataset, 20 \% for the validation dataset, and 20 \% for the test dataset with a distribution of 9573, 3184, and 3192 of videos respectively.

\section{Methods}\label{Theory}
\subsection{Convolutional Neural Networks}
Convolutional Neural Networks (CNNs) are a type of neural networks specifically designed for data with a grid-like topology such as images and videos (\cite{Goodfellow-et-al-2016}). A CNN models this data by using convolutions instead of general matrix multiplications, which means weights are reused on all pixels in each layer. This makes CNNs more computationally efficient than for example a feedforward neural network and less prone to overfitting, because weight sharing also acts as a regularization. The idea of CNNs was introduced in 1989 by (\cite{le1989handwritten}) but was popularized later with the introduction of LeNet-5 (\cite{lecun1998gradient}).

\subsection{SlowFast}
\begin{figure}[h!]
    \centering
    \includegraphics[width=0.95\linewidth]{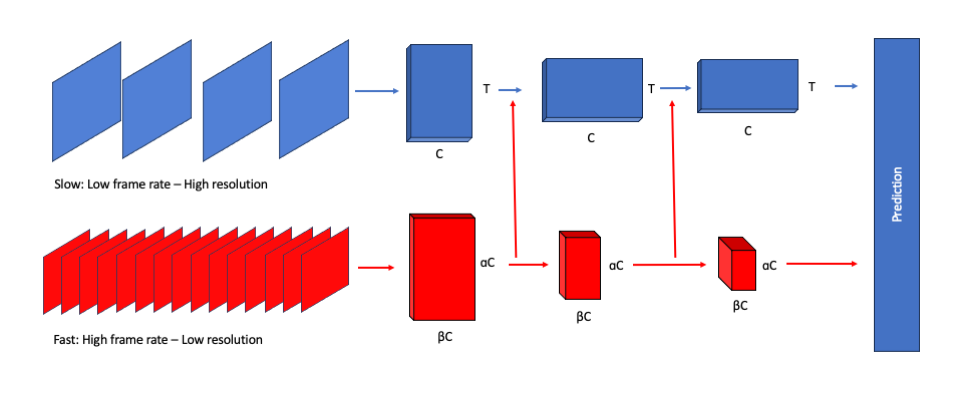}
    \caption{SlowFast architecture. Illustration inspired by  (\cite{feichtenhofer2019slowfast}).}
\label{fig:slowfast_architecture}
\end{figure}

SlowFast is a two-stream network for video analysis tasks presented by (\cite{feichtenhofer2019slowfast}). The core idea is to have two parallel streams working on the input video, one for capturing spatial information and one for capturing temporal information. Both streams consist of inflated ResNet structures, with kernel sizes specifically made to suit the purpose of each stream. A diagram of the overall architecture can be seen in Figure \ref{fig:slowfast_architecture}.

In the following, the network architecture is described in detail. Through an example, the network architecture will be described together with the process of giving a raw video as input to the model and receiving a prediction as output.

A raw video clip of arbitrary image size and length is given as input. The raw clip is scaled to 256 on the shortest side, spatially cropped to $224\times224$, and clipped to a length of 64 frames to ensure a fixed input size is given to the network. The first layer of the network is a ‘data layer’ where each pathway samples a given number of frames from the input. The number sampled depends on the specific architecture of the SlowFast model chosen. For this example, the 4x16 model will be used. The name will be explained later in this example.

\paragraph{Slow pathway}
The slow pathway operates with a large temporal stride. The slow pathway samples frames from the input with a stride of $\tau=16$ resulting in a clip of 4 frames, thereof the name 4x16. The intuition behind the large temporal stride is that the categorical spatial semantics, such as a ‘person’ performing a specific action, does not change their identity throughout the action performed. Therefore, fewer frames should be necessary to model the spatial information in the video. The input clip is run through an inflated ResNet structure. Throughout the network layers, a large number of channels are used to model the high-level features of the spatial semantics. 
%The structure of the slow pathway is similar to the original ResNet architecture, shown previously in Figure \ref{fig:resnet_structures}.
 
\paragraph{Fast pathway}
The fast pathway samples frames with a high frame rate. A parameter $\alpha$ denotes the frame difference between the slow and the fast pathway. Here, $\alpha=8$ because the fast pathway samples 8 times as many frames as the slow pathway. 
The input clip is taken through an inflated ResNet structure. The temporal dimension is kept at a constant size throughout the network layers to avoid losing any temporal information by downsampling. In turn, the fast convolutional network is kept lightweight by limiting the number of channels in each layer. The philosophy is that a complex spatial representation is already provided in the slow pathway, and therefore a high number of channels is redundant in the fast path. A parameter $\beta$ denotes the channel ratio between the two pathways. A value of $\beta = 1/8$ means the fast pathway has 8 times fewer channels than the slow pathway. The authors (\cite{feichtenhofer2019slowfast}) found that the fast pathway operates optimally with 6 to 8 times fewer channels than its slow counterpart. 

\paragraph{Lateral connections}
Between each ‘stage’ of the ResNet architecture, information is fused between the two pathways. This is obtained using ‘lateral connections,’ a technique also used in previous optical flow-based methods (\cite{christoph2016spatiotemporal}, \cite{feichtenhofer2016convolutional}). Unidirectional connections are used to fuse information from the fast pathway to the slow pathway. Since the temporal dimension of the two pathways is not the same, the features of the fast pathway are transformed before fusing. The features are transformed by performing a 3D convolution with a kernel of size (1x1x5), responsible for downsampling the temporal dimension. The transformed features are fused into the Slow pathway by concatenation or summation. 

In the final layer of the ResNet, the slow pathway has an output dimension of (7x7x4), and the fast pathway has an output dimension of (7x7x32). Global average pooling is performed on each output to form two pooled feature vectors. The two feature vectors are concatenated to form the input of the final fully connected classification layer. 
The network design is generic, and the two pathways can be implemented from different types of CNNs. The models used in this paper are constructed on a ResNet 50 (R50), with architectures of 2x32, 4x16 and 8x16. %8x8,

\subsection{Performance evaluation}
This section presents different measures and methods for evaluating the performance of a model. Some methods are described for the binary classification task, but can easily be extended to more classes.

\subsection{Performance measure}
The performance measure is a quantitative measure of a model’s performance. Ideally, one measure should be chosen for direct comparison (\cite{ng2016nuts}). For classification tasks, a common performance measure is accuracy, which measures the proportion of correctly classified examples (\cite{Goodfellow-et-al-2016}). Accuracy is calculated by

\begin{equation}
    \text{Accuracy} = \frac{TP + TN}{TP+FP+FN+TN} \times 100\%%{N}
\end{equation}

Where TP is true positive, TN is true negative, FP is false positive and FN is false negative. Accuracy is the most common performance measure to use for single-label classification. Another common performance measure for classification is the error rate, which can be seen as the opposite of accuracy. The error rate measures the proportion of incorrectly classified examples (\cite{Goodfellow-et-al-2016}). The error rate is calculated by

\begin{equation}
    \text{Error rate} = \frac{FP + FN}{TP+FP+FN+TN} \times 100\% %{N}
\end{equation}

%\subsection{Confusion matrix}
%A confusion matrix is a method for visualizing the errors from a prediction task. The predictions are arranged such that columns represent the predicted labels and the rows represent the true labels.%, see Figure \ref{fig:confusion_matrix_binary}. 
%These characteristics mean that the correct predictions are represented along the diagonal from the upper left corner to the bottom right corner. The confusion matrix supplements the performance measure, as it provides insight into which classes  the high or low performance comes from. Especially when working with class imbalanced data, the performance measure can be misleading, because it is skewed towards the performance on the majority classes (\cite{herlau2016introduction}).
%
%\begin{figure}[h!]
%    \centering
%    \includegraphics[width=0.5\linewidth]{fig4_confusion_matrix.png}
%    \caption{The confusion matrix of a binary classification problem}
%    \label{fig:confusion_matrix_binary}
%\end{figure}

\subsection{Error analysis}
Error analysis is the process of manually looking at the errors from the validation set and analysing where they come from. Error analysis can give clues for the direction of improvement. Different methods can be used for analysing the errors, including the confusion matrix and the size off the diagonal elements. However, in this section the process of Error Analysis as described in (\cite{ng2016nuts}) is presented. The method can be broken down into the following concrete steps:

\begin{enumerate}
    \item  Write down initial ideas for improving the system (in terms of errors to fix). 
    \item Analyse ~$100$ misclassified examples manually and write the result in a spreadsheet containing the categories and comments on the specific examples.%, as seen in Figure \ref{fig:error_analysis_process}.
    \item Write down more categories if new ones occur while looking at the examples.
    \item Calculate the percentage of errors each category accounts for i.e., the percentage of errors that can be eliminated at most if that category's error source is addressed.
    \item Decide which error to work on based on highest error elimination potential, how much progress you expect to make and how much work it takes.
\end{enumerate}

The process is iterative and the errors can be revisited whenever adjustments have been made to the system.

%\begin{figure}
%    \centering
%    \includegraphics[width=1.0\linewidth]{fig5_error_analysis.png}
%    \caption{The process of error analysis elaborated by Andrew Ng. Image from (\cite{ng2016nuts}). Analyse 100 misclassified examples manually and write the result in a spreadsheet
%containing the categories and comments on the specific examples.}
%    \label{fig:error_analysis_process}
%\end{figure}

\section{Results}\label{Results}
\subsection{Training}

The training curves are shown to highlight the training and validation performance, and to show whether avoiding over fitting was successful, and whether the models were trained to end or if more could be learned.

The THETIS data only contains 1980 videos, making it fairly fast to train. For the data set, dropout, data augmentation, early stopping, and weight decay were used to avoid over fitting. 

% Insert table of training times on Nvidia gpu

% Show figure of the training validation error curves for 4x16 on thetis
\begin{figure}[h!]
    \centering
    \includegraphics[width=0.7\linewidth]{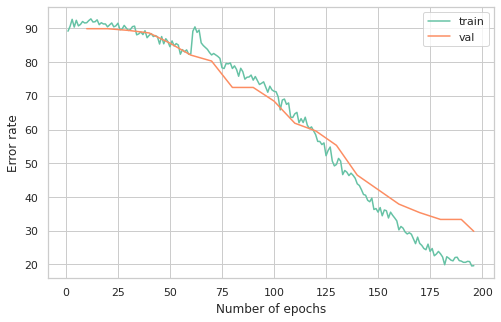}
    \caption{4x16 models training curve on the THETIS Data-set}
    \label{fig:train_val_error_thetis_4x16}
\end{figure}

For the SlowFast 4x16 model, a plot showing the training and validation error on the THETIS data set can be seen in Figure \ref{fig:train_val_error_thetis_4x16}. From the figure, it is clear that the training and validation errors follow each other consistently until roughly epoch 125, where the gap between them increases more and more, indicating that the model has over-fitted to the training data. Clearly, we can avoid this over fitting using early stopping. However, since the validation error keeps decreasing we are still learning generalized information, and stopping earlier would mean important information would get lost. Therefore, a training time of 196 epochs are used because it results in the minimum validation error. Both error curves are still decreasing at epochs 196 although they have flattened out slightly, indicating that more can potentially be learned.

\subsection{Comparing model performances}
% Compare and evaluate the results of the different model architectures trained
The different models were trained on the data set to assess which model is most suitable for the task of classifying tennis videos. In Table \ref{tab:model architecture comparisson} the training and validation accuracy of the different models are shown for the THETIS dataset.

\begin{table}[]
\caption{Different R50 SlowFast architectures trained on the THETIS data. For validation, one random clip is sampled. For testing, three spatial crops from two temporal clippings are sampled and the predictions are summed to make get the final prediction. The accuracies are given in \%.}
\centering
\begin{tabular}{l|l|l|l|l|}
\cline{2-5}
                                                   & Architecture       & train acc & val acc & 3x2 ens. test acc \\ \hline
\multicolumn{1}{|l|}{Models trained on THETIS data} & SlowFast 2x32, R50 & 9.63     & 10.10   & 7.29             \\ \cline{2-5} 
\multicolumn{1}{|l|}{}                             & SlowFast 4x16, R50 & 80.39     & 70.20   & 73.96             \\ \cline{2-5} 
\multicolumn{1}{|l|}{}                             & SlowFast 8x8, R50  & 83.98     & 67.17   & 71.88             \\ \hline
\end{tabular}
\label{tab:model architecture comparisson}
\end{table}

\subsection{The best model}
The results are shown in table \ref{tab:model architecture comparisson}, and the SlowFast 4x16 model performs better on the THETIS dataset with $3.03\%$ higher accuracy on the validation set than the Slowfast 8x8 model. The SlowFast 2x32 model has the lowest accuracy on both datasets, with an extremely low accuracy on the THETIS dataset relative to the two other models. Due to the low acccuracy of the 2x32 model, this model is not considered a valid candidate in the following sections. 

\subsection{Analysing the errors}
In this section the results of the methods used for error analysis will be presented. Two methods were used for analysing the errors. First, the distribution of true and false predictions among the classes were analysed through a confusion matrix. Second, the falsely labelled videos were categorized according to the error analysis method (\cite{ng2016nuts}). When categorizing errors it is important to keep in mind, that the error categories represent categories identified when looking at the videos manually. The category is observed in the given error, but is not necessarily the reason for the error.

% Thetis - shots
\subsection{THETIS: Classifying types of tennis shots}
\begin{figure}
    \centering
    \includegraphics[width=0.85\linewidth]{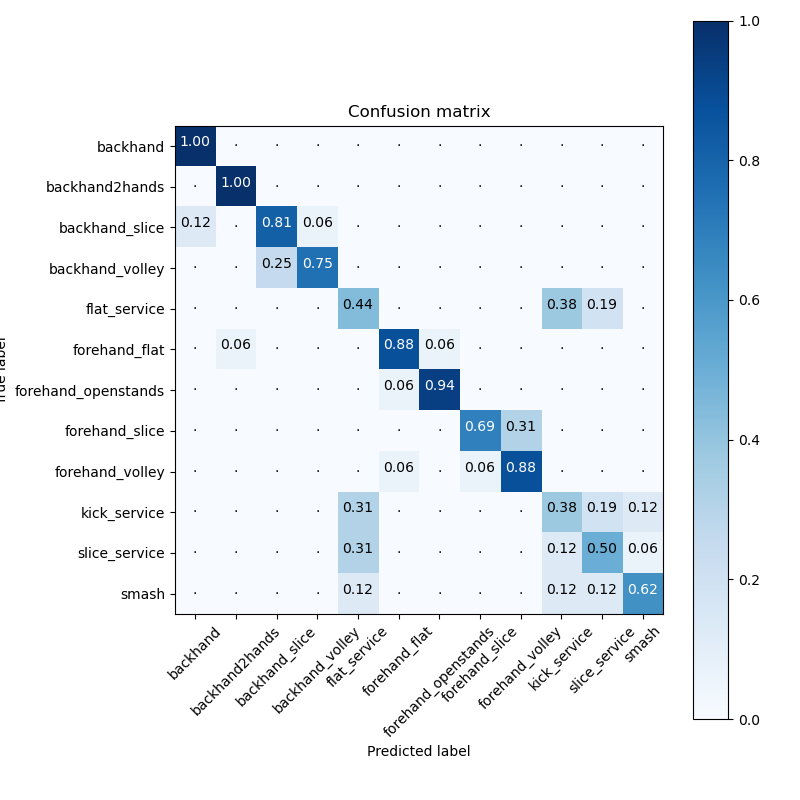}
    \caption{Confusion matrix showing the performance of SlowFast 4x16 on the THETIS data test set}
    \label{fig:confusion_matrix_thetis_4x16}
\end{figure}

The true and false predictions of the SlowFast 4x16 model on the THETIS test set were visualized in a confusion matrix, see Figure \ref{fig:confusion_matrix_thetis_4x16}. From the confusion matrix it is clear, that the accuracy varies throughout the different classes all the way from $38\%$ to $100\%$. Considering the overall accuracy was found to be $73.96 \%$, overall accuracy clearly gives a simplified idea of the model performance, that could easily be misinterpreted. 
The confusion matrix shows, that the easiest classes to classify are backhand and backhand2hands while the most difficult classes are flat service and slice service. It is also clear, that many of the classes that belong to the same overall class such as serve and forehand are confused with each other.

%Confusion matrix thetis test 4x16

\subsubsection{Looking at misclassified samples}
To understand why these classes are more easy/difficult to classify, the misclassified samples were examined using error analysis. The error categories identified and the corresponding amount of errors (\%) are shown in Table \ref{tab:error_analysis_thetis}. The alert reader may have noticed that the error categories exceed $100 \%$. The reason being that an error can belong to multiple categories. For example, it is possible that an error belong to a confusion between smash and serve, but because the player is a beginner, he might have forgotten to lift his arm. 

%Error analysis table thetis - lav evt bedre kategorier
\begin{table}
\centering
\caption{Error analysis of the SlowFast 4x16 model on the THETIS dataset. 
}
\begin{tabular}{ll|l}
& error category & amount (\%) \\ \hline
{$\left.\begin{array}{l}
                \text{no ball}\\
                \end{array}\right\lbrace$} & serve confusion    &         $44.4$   \\
 & slice/volley confusion   &        $20.4$     \\
& smash/serve confusion       &   $16.7$          \\
& beginners         &   $9.3$ \\
& others        & $14.8$ \\
\end{tabular}
\label{tab:error_analysis_thetis}
\end{table}

% Error analysis table - explain and comment on the different categories and which solutions exist for each of them. 

The error analysis illustrates that most of the errors come from serves being confused with each other. In Section \ref{Chap:data}, the technical differences between the serves were described. Here it was clear that the large movements of the serves are very similar and they differ by small movements like racquet angle and motion direction when the racquet hits the ball. These small movements creates different spin on the ball resulting in very different trajectories. However, without a ball in the videos the differences may be too small to detect.

Similarly, other confusions are likely caused by the shots looking very similar without a ball. For example, smash and serves are often confused, because the lifted arm of the smash can easily be confused with the throwing arm of the serve, when no ball is present. It also applies to the slice and volley shots, mostly for the forehand categories which account for $16.7 \%$ of the total slice/volley confusion errors. Again, these movements are similar without a ball because the grip, angle of the racquet, and direction of movement are alike. Normally, the volley is played closer to the net, before the ball hits the ground, meaning additional cues such as position on the court are also missing in distinguishing the two categories slice and volley. $9.3 \%$ of the errors are classified as beginners. This category was used to describe the shots performed more sloppy. %Impossible for the human eye to recognize.

\section{Conclusion and discussion}
The aim of this work was to investigate to which degree deep learning, in particular the SlowFast architecture, can be used to classify actions in tennis videos and to identify the associated challenges. 

By studying the recent advances in video classification and tennis video classification, the methodological foundation for working with deep learning models in the field was laid using the state of the are framework PySlowFast for video classification of the THETIS RGB data-set. Three selected SlowFast architectures were trained and evaluated on the the THETIS RGB data-set. The model performances were evaluated and compared based on accuracy as primary performance measure and inference time as secondary performance measure. The two performance measures were used to outweigh the advantages and disadvantages of the models in relation to the application. By analysing the errors through plots and manually looking at misclassified samples, a deeper understanding of the modelling task was obtained. Throughout the entire process potential challenges of classifying actions in tennis videos using deep learning were identified. The results showed that we are able to classify tennis actions to a large degree, with accuracy exceeding the results of existing work. The SlowFast 4x16 model had the highest generalization accuracy on the THETIS RGB data-set with a 74 \% a big improvement as compared to the slightly different statistics of a F1 at 47\% in (\cite{vinyes2017deep}). Overall, the nature of the SlowFast architecture with a spatial and a temporal path showed promising results for tennis data.

The network only exhibited smaller challenges, which we suspect are caused when distinguishing classes where only small spatio-temporal cues differentiate them, such as racquet angle or arm and racquet movement relative to the body, and classes where the cues differentiating them take up a small portion of the frame. For example, the volley and the slice shots are nearly indistinguishable in the THETIS RGB data-set based on movement and the lack of player position on the court and absence of a ball mean that important information to distinguish such cases are missing. A set of potential challenges were also identified through error analysis. The limitations of the THETIS RGB data-set is that it is not collected from a tennis court and no ball was used also mentioned in (\cite{vinyes2017deep}). This lacks both the trajectory of the ball and the actual position on the court while performing the action, this makes the dataset mainly viable for academic purposes and can at most be used for transfer learning to help initialize weights in video recognition tasks with respect to real tennis matches.

The identification of these potential challenges and limitations can be helpful in the future by highlighting the potential limitations to beware of and suggesting ways of improving the THETIS RGB data-set. 

For the further progress of tennis research, high quality data-sets are needed with high resolution recordings from real tennis matches with time stamps of different events as scores and tracking information as player position, ball position and bounce positions. Further research would then be possible with respect to both tracking research and statistical analysis of shots in tennis matches.  

%Bibliography
\newpage

%\bibliographystyle{unsrt} 
%\bibliographystyle{bibtex}
%\bibliography{references.bib}
%\section{References}
%\renewcommand\refname{ }
%\bibliographystyle{unsrt}
%\bibliographystyle{apalike}
%\bibliographystyle{reading}
%\bibliographystyle{agsm}
%\addcontentsline{toc}{section}{References}

%\bibliographystyle{plain}
%\bibliography{references.bib}

% old reference with biber
\printbibliography[heading=bibintoc]

\end{document}